\def\year{2021}\relax
\documentclass[letterpaper]{article} 
\usepackage{aaai21}  
\usepackage{times}  
\usepackage{helvet} 
\usepackage{courier}  
\usepackage[hyphens]{url}  
\usepackage{graphicx} 
\usepackage{natbib}  
\usepackage{caption} 
\usepackage{graphicx}
\usepackage{xcolor}
\usepackage{dblfloatfix}
\usepackage{capt-of}
\usepackage{algorithm}
\usepackage{algorithmic}
\usepackage{amsmath,amsfonts,amssymb}

\title{Improving Robot-Centric Learning from Demonstration via Personalized Embeddings}
\author {
    Mariah L. Schrum ,\textsuperscript{\rm 1 \thanks{Both authors contributed equally to this research}}
    Erin Hedlund, \textsuperscript{\rm 1  $^{*}$ }
    Matthew C. Gombolay \textsuperscript{\rm 1} \\
}

\affiliations {
    \textsuperscript{\rm 1} Institute for Robotics and Intelligent Machines \\
    Georgia Institute of Technology \\
    Atlanta, GA \\
    
    mschrum3@gatech.edu, ehedlund6@gatech.edu, matthew.gombolay@cc.gatech.edu
}

\newcommand\changeNew[1]{\textcolor{black}{#1}}

\begin{document}

\maketitle

\begin{abstract}
Learning from demonstration (LfD) techniques seek to enable novice users to teach robots novel tasks in the real world. However, prior work has shown that \textcolor{black}{robot-centric} LfD approaches, such as Dataset Aggregation (DAgger), do not perform well with human teachers. DAgger requires a human demonstrator to provide corrective feedback to the learner either in real-time, which can result in degraded performance due to suboptimal human labels, or in a post hoc manner which is time intensive and often not feasible. To address this problem, we present Mutual Information-driven Meta-learning from Demonstration (MIND MELD), which meta-learns a mapping from poor quality human labels to predicted ground truth labels, thereby improving upon the performance of prior LfD approaches for DAgger-based training. The key to our approach for improving upon suboptimal feedback is mutual information maximization via variational inference. Our approach learns a meaningful, personalized embedding via variational inference which informs the mapping from human provided labels to predicted ground truth labels. We demonstrate our framework in a synthetic domain and in a human-subjects experiment, illustrating that our approach improves upon the corrective labels provided by a human demonstrator by 63\%.
\end{abstract}

\section{Introduction}
In Learning from Demonstration (LfD), a robot seeks to perform a task by observing human task demonstrations \cite{argall_survey_2009}. \textcolor{black}{In human-centric LfD, the human demonstrator drives the interaction and provides the demonstrations for each trajectory. In robot-centric LfD, a demonstrator observes the robot and must provide corrections to the robot learner for the learner to gather new information and improve upon its current policy \cite{Laskey2017}.  One such example of human-centric learning is Behavioral Cloning (BC) in which the demonstrator provides a series of demonstrations and the agent is trained via supervised learning. BC, however, suffers from covariate shift and performs poorly when when the environment's transition dynamics are stochastic \cite{Ross2010,Osa2018}. To overcome this limitation, \citet{Ross2011} introduced a robot-centric approach, called Dataset Aggregation (DAgger). DAgger learns a policy, $\pi_0$, from the initial trajectories provided by the demonstrator. $\pi_0$ is then rolled out and the demonstrator provides corrective feedback. The new, corrective labels provided by the demonstrator are aggregated with the previous trajectories and used to train a new policy. }

\textcolor{black}{Prior work in robot-centric LfD has shown that DAgger outperforms human-centric LfD algorithms, such as BC, when the demonstrator is an oracle that provides high-quality labels \cite{Ross2011}. However, such studies do not necessarily translate to the real-world with human demonstrators \cite{Amershi2014,Spencer2020,Berggren2019}}. Prior work by  \citet{Laskey2017} has shown that DAgger performs poorly and may even perform worse than BC when the demonstrator is a human and the learner is a neural network. 
\textcolor{black}{DAgger's poor performance} is due to the fact that humans often provide poor quality feedback \cite{Sena2020}. Furthermore, humans differ in the way they provide this feedback depending on the task and the human's abilities \textcolor{black}{\cite{Sammut1992,Paleja2019}.} This suboptimality and heterogeneity can make it difficult for robots to learn from human teachers. To effectively learn from a human demonstrator, robot-centric LfD approaches must take into account a teacher's demonstration style to correct for the teacher's suboptimality and improve the policy of the learner. Yet, no prior work has investigated correcting for demonstrator suboptimality while accounting for heterogeneity among demonstrators.

To overcome this limitation, we introduce Mutual Information-driven Meta-learning from Demonstration (MIND MELD), which  meta-learns an individual-specific mapping from human labels to predicted ground truth labels via a Long Short-Term Memory (LSTM) based neural network architecture. Because individuals  differ in the way that they provide feedback, we propose to learn a personalized embedding via variational inference that encapsulates information about individual tendencies and corrective styles.  This personalized embedding informs the mapping of an individual's suboptimal labels to labels that more closely approximate the ground truth, thus improving upon the performance of robot-centric LfD methods for DAgger-based training.

To evaluate the ability of MIND MELD to learn meaningful embeddings and improve upon human-provided, suboptimal corrective labels, we conduct a human-subjects study in which we recruit human demonstrators to provide corrective feedback to an agent.  Additionally, we investigate if the learned personalized embeddings capture salient aspects of demonstrator style via correlation analysis between the learned embeddings, stylistic tendencies, personality traits, and experience metrics.

In our work, we contribute the following:

\begin{enumerate}
    \item \textcolor{black}{We create} a novel, personalized learning from demonstration framework, MIND MELD, for inferring individual demonstrator styles and improving upon suboptimal corrective labels.
    \item \textcolor{black}{We conduct} a human-subjects study in which participants provide corrective feedback in a series of tasks to train MIND MELD.
    \item \textcolor{black}{We present} results that demonstrate the ability of  MIND MELD  to improve upon suboptimal human labels and learn meaningful representations of demonstrator style. \textcolor{black}{We show that MIND MELD is able to improve suboptimal human-provided labels by 63\% by inferring personalized embeddings. We demonstrate that these embeddings significantly correlate with stylistic tendencies of the demonstrator ($p<.001$).}
\end{enumerate}

\begin{figure*}[h!]
\centering
\includegraphics[width=\linewidth]{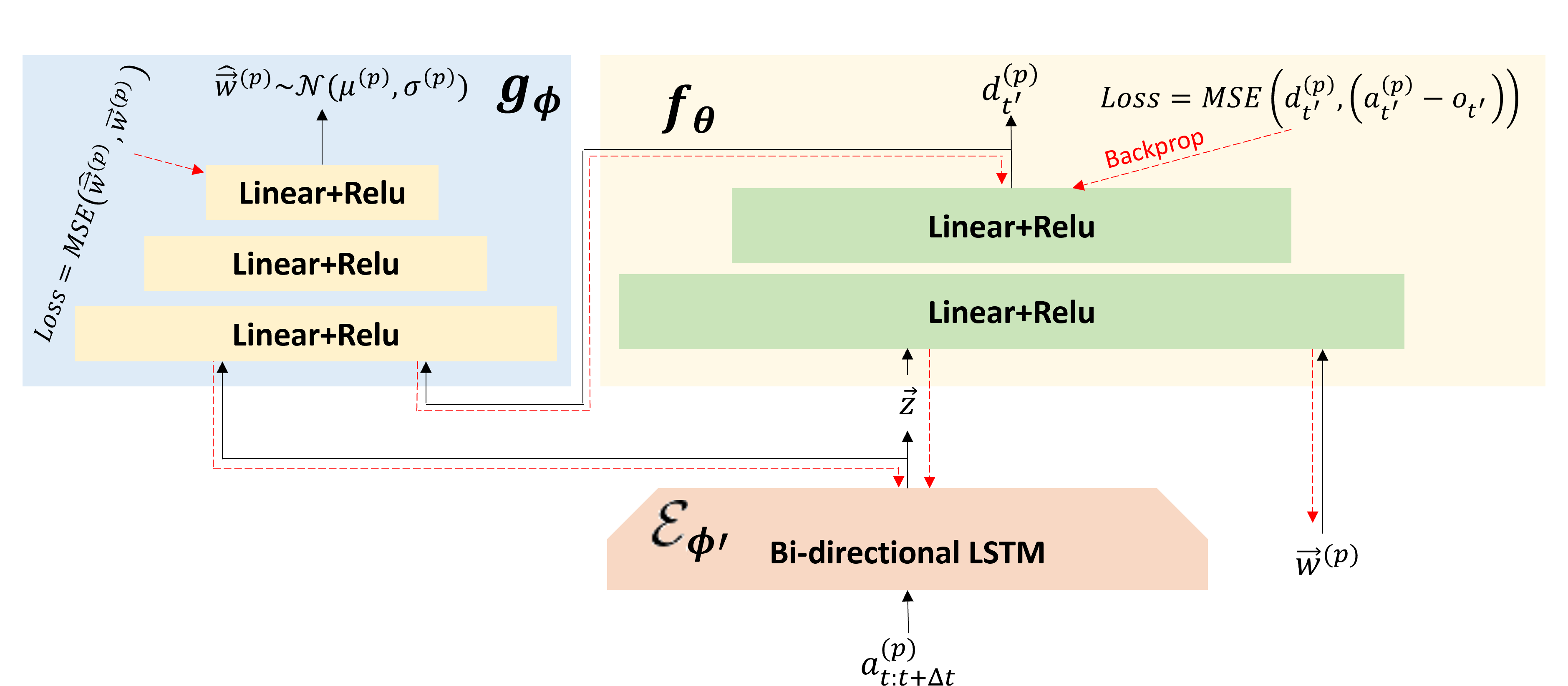}
\captionof{figure}{This figure shows the network architecture. $a_{t'}^{(p)}$ represents a demonstrator, $p$'s, corrective label, $a$, at time, $t'$. $g_{\phi}$ serves to maximize mutual information between the encoding $\vec{z}$, the output $d_{t'}^{(p)}$ and the learned embedding $\vec{w}^{(p)}$. The objective is to minimize the mean squared error, (MSE), between the predicted difference, $d_{t'}^{(p)}$, and the true difference, $a^{(p)}_{t'}-o_{t'}$ , of the demonstrator's corrective feedback and the ground truth label, $o_{t'}$. We feed the sequence of corrective feedback, \textcolor{black}{$a_{(t:t+\Delta t)}^{(p)}$,} from time $t$ to \textcolor{black}{$t+\Delta t$,} into the bi-directional LSTM to extract sequential information that may be informative for predicting the ground truth label at time, $t'$. The LSTM has a hidden layer with 32 neurons. $f_{\theta}$ has a hidden layer consisting of 200 neurons and $g_\phi$ has two hidden layers with 64 and 32 neurons respectively.}
\label{fig:NetworkStructure}
\end{figure*}

\section{Related Works}
Prior work has investigated supervised learning approaches to learn a mapping function from states to actions based on trajectories provided by an expert demonstrator \cite{Chernova2009,Ross2011, LFDReview,Liu}. The problem encountered when learning this mapping is that the independent and identically distributed (i.i.d.) assumption is violated because the learner’s predictions affect future states \cite{Ross2010, Jena} and the number of mistakes the learner makes is quadratically proportional to the time horizon, $T$.  

To address this problem, Ross et al. introduced Dataset Aggregation (DAgger) which aggregates a training set based on expert labels queried during a policy rollout instead of relying on a mixture of previous policies \cite{Ross2011}. DAgger utilizes the state distribution induced by the current policy to request labels from the expert and a gating function determines the mixture of expert and learner during each rollout. The authors of this work prove similar linear loss guarantees to prior work and show that DAgger empirically outperforms prior work. 

While DAgger performs well when quality expert labels are provided, \citet{Laskey2017} demonstrated that robot-centric learning approaches such as DAgger can result in mislabelling and therefore poor performance of the learner.  The authors show that human-centric learning, in which the expert demonstrates the task to the learner, can actually outperform DAgger.
Additionally, DAgger suffers from the high work load it places on the demonstrator which can result in expert fatigue and poor results \cite{HG-Dagger, Laskey2016, Packard2018}. \textcolor{black}{Furthermore, it can be impractical for humans to provide corrective feedback to DAgger in real-time \cite{ross_learning_2013}.}

To overcome this challenge posed by robot-centric LfD, prior work has attempted to relieve the burden placed on the expert by DAgger and improve upon the interaction with the learner \cite{Coaching,HG-Dagger,Spencer2020,Menda2019}. Daume et al. proposed an imitation learning by coaching algorithm in which the coach gradually provides more and more difficult actions for the learner to imitate \cite{Coaching}. The coach demonstrates actions that are preferred by the learner, meaning they induce a lower task loss. Results show that such a coaching scheme is able to outperform DAgger and achieve a lower regret bound.  To reduce the workload of the expert and improve upon the provided demonstrations, Kelly et al. proposed HG-DAgger which allows the expert to operate the gating function, meaning that the expert decides when to be in control and when to passively observe the learner \cite{HG-Dagger}. This alleviates the difficulty of providing accurate labels and consequently learns a stationary policy which stabilizes around the expert trajectories. Spencer et al. expands on this idea and utilizes both expert interventions and non-interventions to learn a policy in the Expert Intervention Learning (EIL) algorithm \cite{Spencer2020}. 

\color{black}
While prior work has learned LfD policies from heterogeneous demonstrators \cite{paleja_interpretable_2021} and attempted to improve upon DAgger by handing control back to the human \cite{HG-Dagger}, our approach is the first to  improve upon robot-centric learning by inferring demonstrator style.  Additionally, there is a need for LfD algorithms that can effectively learn from suboptimal and heterogeneous demonstrators \cite{LFDReview}.  \color{black}



 \begin{figure*}
     \centering
     \includegraphics[width=\linewidth]{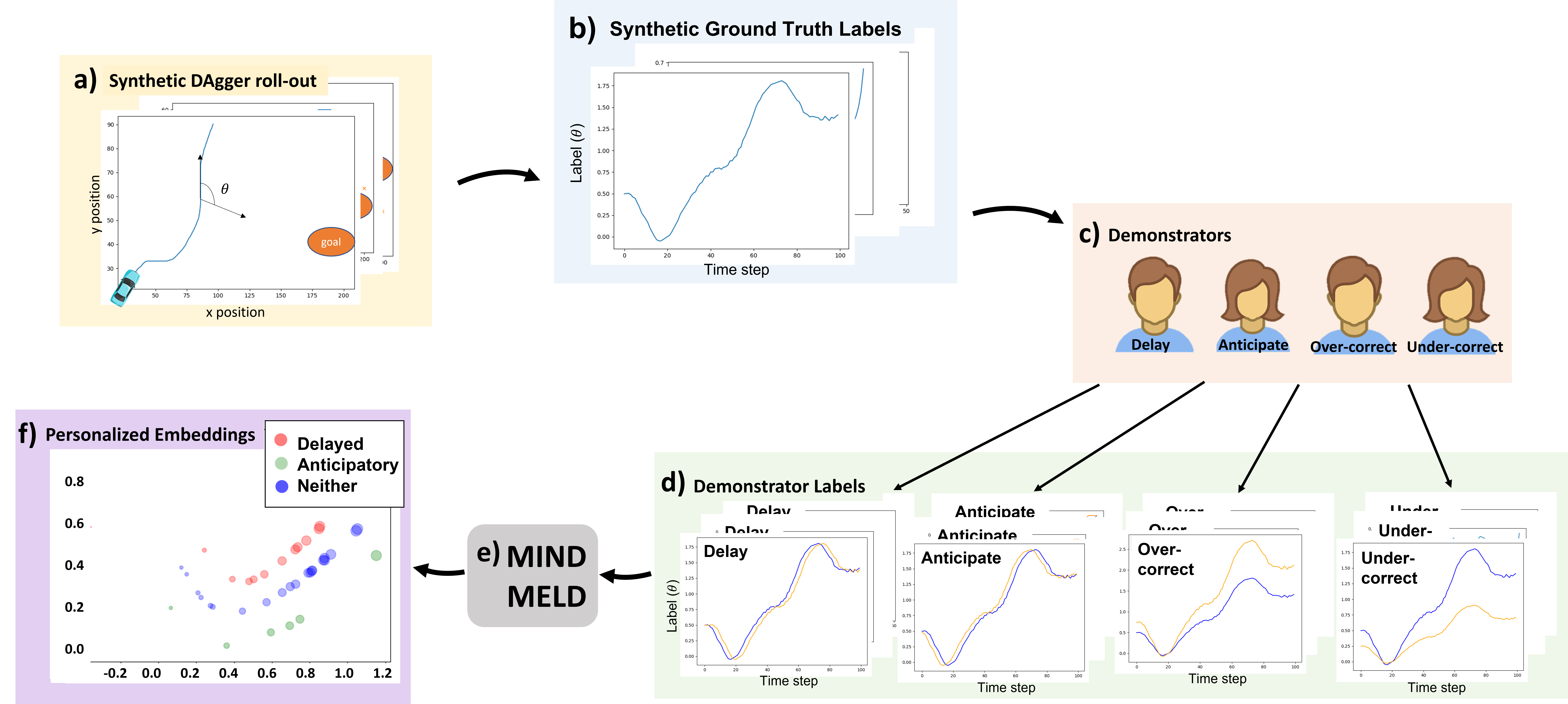}
   \caption{This figure shows the creation of the synthetic data. a) shows the artificial DAgger rollouts, b) the ground truth labels, c) the demonstrators, and d) the corrective feedback.  The mapping of suboptimal labels via our architecture, MIND MELD, produces embeddings shown in f). In f), the diameter of a point represents the degree to which an individual over- or under-corrects. The color represents the individual's style (i.e., delayed, anticipatory, or neither).}
     \label{fig:synthetic_data}
 \end{figure*}

\section{Methodology}
\label{sec:methods}
In this section, we discuss our methodology for improving upon robot-centric learning from demonstration. We first outline our network architecture and learning scheme for personalized embeddings and label mapping. We next describe our simulated  experiment  for  proof-of-concept  and  to  verify  our network architecture in \textcolor{black}{Section \ref{sec:synthetic}.} Lastly, we describe our human-subjects experiment with humans as demonstrators in \textcolor{black}{Section \ref{sec:humanexperiment}.}

\color{black}
\subsection{Preliminaries}
We frame the LfD problem as a Markov Decision Process \textcolor{black}{sans reward function (MDP$\backslash$R). The MDP$\backslash$R  is defined by the 4-tuple $\langle\mathcal{S},\mathcal{A},\mathcal{T},\gamma\rangle$.} $\mathcal{S}$ represents the set of states and $\mathcal{A}$ the set of actions. $T: \mathcal{S} \times \mathcal{A} \times \mathcal{S}' \rightarrow [0,1]$ is the transition function that returns the probability of transitioning to state $s'$ from state, $s$, applying action, $a$.  $\gamma$ weights the discounting of future rewards. Reinforcement learning seeks to synthesize a policy, $\pi: \mathcal{S} \rightarrow \mathcal{A}$, mapping states to actions to maximize future expected reward. In an LfD paradigm, a demonstrator provides a set of trajectories, $\{(s_t,a_t), \forall t \in \{1,2,...T\}\}$, from which the agent learns a policy.
\color{black}

\subsection{Assumptions}
In our methodology, we make the following assumptions.

\begin{itemize}
\item In robot-centric learning from demonstration tasks, humans provide corrective feedback that is suboptimal \textcolor{black}{with respect to an optimal planner}.
  \item These suboptimal strategies are \textcolor{black}{identifiable} and  can be represented as ``styles'' via a learned embedding.
  \item Humans provide corrective feedback in a \textcolor{black}{predictable} way across different tasks.
 
\end{itemize}

Based on these assumptions, we can learn an individual's corrective style to inform a mapping from suboptimal labels to ground truth labels.

\color{black}
\subsection{Architecture}
 Fig. \ref{fig:NetworkStructure} shows our network architecture. Our architecture consists of three components: 1) The bidirectional LSTM encoder, $\mathcal{E}_{\phi'}\colon A \rightarrow Z$, 2) the prediction subnetwork, $f_\theta\colon Z \times W \rightarrow \mathbb{R}$, and 3) the mutual information subnetwork, $g_\phi\colon Z \times \mathbb{R} \rightarrow \mathcal{N}_W$. The label that we are trying to improve upon is designated as $a_{t'}^{(p)}$ for demonstrator, $p$, where \textcolor{black}{$t' \in (t,t+\Delta t)$.} $Z \subset \mathbb{R}^{k}$ is the set of k-dimensional encodings extracted from the sequence of corrective feedback.  We utilize the bi-directional LSTM, $\mathcal{E}_{\phi'}$, to extract an encoding, $\vec{z} \in Z$, for the sequence of corrective labels, \textcolor{black}{$a_{(t:t+\Delta t)}^{(p)}$, provided from time $t$ to $t+\Delta t$ by person $p$.}
 
 $W=\{\vec{w}^{1}, \vec{w}^{2}...\vec{w}^{P}\} \subset \mathbb{R}^d$ is the set of d-dimensional personalized embeddings.  $f_\theta$ maps the encoding, $\vec{z}$, and personalized embedding, $\vec{w}^{(p)}$, to the predicted difference, $d_{t'}^{(p)} \in \mathbb{R}$, between the ground truth label, $o_{t'}$, and the individual's corrective label, $a_{t'}^{(p)}$. \textcolor{black}{Because we train $g_\phi$ to maximize the log-likelihood as discussed in the next section, this subnetwork recovers a multivariate normal distribution, $\mathcal{N}_W$ \cite{Paleja2019}}. Specifically, this subnetwork learns a mapping of the encoding, $\vec{z}$, and predicted difference, $d_{t'}^{(p)}$ to a normal distribution (representing the posterior distribution) of the demonstrator's personalized embedding,  $\vec{w}^{(p)}$. \changeNew{We initialize $w^{(p)}$ based upon the prior, $\hat{w}^{(p)} \sim \mathcal{N}(0,1)$, and} sample from the approximate posterior to produce $\hat{\vec{w}}^{(p)}$, representing an estimate of the embedding.
 \color{black}

\subsection{Variational Inference} In this work, we are motivated by the assumption that humans exhibit various and distinct styles when providing corrective labels to the learner. Therefore, to accurately correct for suboptimal demonstrations, we encapsulate information about the individual's corrective style via a personalized embedding, $\vec{w}^{(p)}$, for individual, $p$, as described in Eq. \ref{eq:mutualinfo}. In our formulation, we want to maximize mutual information between  our learned personalized embedding, $\vec{w}^{(p)}$,  the encoding of the demonstrator labels, $\vec{z}$, and corrective mapping, $d_{t'}^{(p)}$, to ensure that our embeddings are capturing the salient information about the demonstrator's style.  Intuitively, maximizing mutual information means that observing informative corrective feedback should reduce the uncertainty of our learned embedding. 

Because maximizing the mutual information requires access to an intractable posterior distribution, $P(\vec{w}^{(p)}|\vec{z},d_{t'}^{(p)})$, we  employ variational inference and the evidence lower bound to reach a solution as shown in Eq. \ref{eq:mutualinfo}. \textcolor{black}{Further details on the derivations can be found in  \citet{Chen2016}}.  Via this formulation, we thus encourage $\vec{w}$ to encapsulate salient information about the demonstrator's style.  The mutual information between $\vec{z}$,  $d_{t'}^{(p)}$ and personalized embedding, $\vec{w}^{(p)}$, is denoted  as $I(\vec{w}^{(p)};\vec{z},d_{t'}^{(p)})$. 
The variational lower bound is $L_I(f_{\theta|\vec{w}},g_{\phi|\theta})$ .

\begin{align}
    & I(\vec{w}^{(p)};\vec{z},d_{t'}^{(p)})=H(\vec{w}^{(p)})-H(\vec{w}^{(p)}|\vec{z},d_{t'}^{(p)})\geq \label{eq:mutualinfo}
    \\& \mathop{\mathbb{E}}[log(g_\phi(\vec{w}^{(p)}|\vec{z},d_{t'}^{(p)}))]+H(\vec{w}^{(p)}) = L_I(f_{\theta|\vec{w}},g_{\phi|\theta})
     \nonumber
\end{align}

We utilize two separate loss functions to train our network to learn both the embedding, $\vec{w}^{(p)}$ and the difference, $d_{t'}^{(p)}$ as shown in Fig. \ref{fig:NetworkStructure}. We minimize the mean squared error between the sampled embedding approximation, $\hat{\vec{w}}^{(p)}$, and the personalized embedding, $\vec{w}^{(p)}$, \textcolor{black}{which is equivalent to maximizing the log-likelihood of the posterior.} We also minimize the mean squared error between  $d_{t'}^{(p)}$ and the difference between the ground truth embedding, $o_{t'}$ and $a_{t'}^{(p)}$. These losses are summed (Eq. \ref{eq:loss}), and are backpropagated through the layers and the input embedding, $\vec{w}^{(p)}$, to update and learn the embedding during training. Therefore, during training, the personalized embedding  will eventually converge to  reflect the individual's feedback style. During test time, this personalized embedding informs the mapping of new feedback to improved labels.

\begin{align}
   & L_{\theta,\phi,\phi',\vec{w}}=\frac{1}{K+1}\sum_{k=0}^{K}\Big((\hat{\vec{w}}^{(p)}_k-\vec{w}^{(p)}_k)^2+\nonumber
    \\& \indent\indent\indent\indent\indent\indent
 (d_k^{(p)}-(a_k^{(p)}-o_k))^2\Big)
    \label{eq:loss}
\end{align}

 \section{Synthetic Experiment}
\label{sec:synthetic}

\textcolor{black}{Before evaluating MIND MELD on human demonstrators, we first conduct a synthetic experiment to fine-tune the architecture and evaluate MIND MELD's ability to correct for suboptimal labels. To do so, we simulate a driving task in which the objective of the demonstrator is to teach the agent to drive to a goal location in the environment. The state space is defined as the position of the car and the continuous  action space is defined as the angle of the wheel. We assume that turning the wheel $\Delta\theta$ causes the car to turn by an equivalent amount.}  

To create synthetic training data by which to train our architecture, we first create a set of artificial DAgger-like roll-outs (Fig. \ref{fig:synthetic_data}a). The ground truth labels are calculated as the difference in the heading of the agent and the angle to the goal (Fig. \ref{fig:synthetic_data}b). We create a set of artificial, suboptimal demonstrators by randomly assigning each demonstrator either a delayed (actions are executed later in time compared to the ground truth), anticipatory (actions are executed  sooner in time compared to the ground truth), or neither style (actions temporaly match the ground truth) and to be either an over-corrector (actions are greater in magnitude compared to the ground truth) or under-corrector (actions are smaller in magnitude compared to the ground truth) by a randomly selected magnitude (Fig. \ref{fig:synthetic_data}c). This ``style'' is then utilized to map the ground truth labels to suboptimal, artificial human labels (Fig. \ref{fig:synthetic_data}d). We employ this artificial data to demonstrate the ability of our architecture to correct for poor human labels. 

\subsection{Results}
\textcolor{black}{Fig. \ref{fig:synthetic_data}f shows the results of the learned embeddings  plotted in latent space.
 The} embeddings for individuals that greatly over-correct are clustered towards the right of the graph and those that greatly under-correct are located towards the left. Those who neither over-correct nor under-correct are located at the elbow in the plot. Additionally, those who provided delayed feedback are located towards the top of the plot and those who provided anticipatory feedback are located towards the bottom. These results confirm that our embeddings learn  meaningful representations of an individual's feedback style. Furthermore, we confirm that our architecture successfully maps the suboptimal feedback to feedback that is closer to the ground truth embeddings. We find a 61\% improvement of labels in the calibration tasks. For unseen test tasks that are not used to train our network, we find a 55\% improvement in the quality of the labels after mapping. These results show that MIND MELD is able to learn meaningful personalized embeddings and utilize these embeddings to improve upon suboptimal corrective feedback.

\begin{figure}[h]
    \centering
    \includegraphics[width=.7\columnwidth]{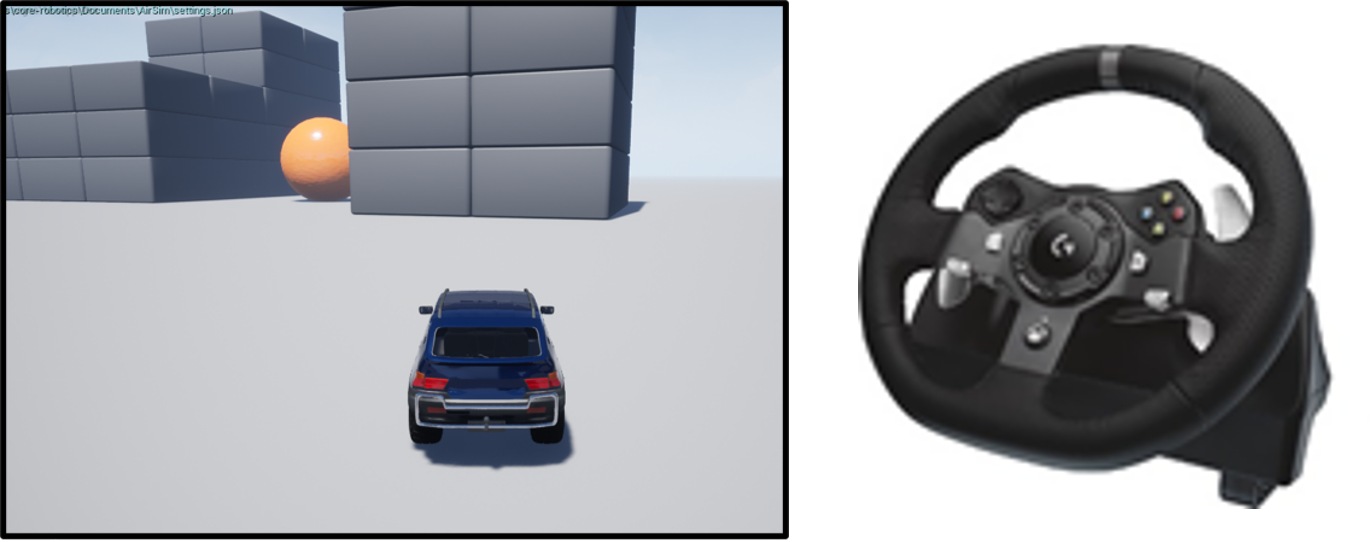}
  \caption{This figure shows the simulator and steering wheel utilized in our human-subjects experiment.}
    \label{fig:simulator}
\end{figure}

\section{Human-Subjects Experiment}
\label{sec:humanexperiment}
We  conduct a human-subjects experiment to evaluate our architecture with human demonstrators and illustrate MIND MELD's ability to improve upon suboptimal corrective feedback. Our study has been approved by the IRB under protocol H19630.

\subsection{Driving Simulator Domain}
We evaluate MIND MELD in a driving simulation domain with human demonstrators, \textcolor{black}{a common task in prior LfD work \cite{Laskey2016, Ross2011}}.  We employ the AirSim driving simulator based on Unreal Engine in conjunction with an Xbox steering wheel and pedals, shown in Fig. \ref{fig:simulator}. We utilize a simple Blocks environment in which the objective of the LfD task is to drive to a large orange ball. \textcolor{black}{The state space is defined as the position, body velocity, body acceleration of the car, and the image provided by the camera located on the front of the car. The action space is the position of the wheel and is constrained to be between -2.5 to 2.5.}

We create a series of twelve synthetic, DAgger-like rollouts, an example of which is shown in Fig. \ref{fig:rrt}. These rollouts are  representative examples of DAgger rollouts that allow us to capture the feedback styles of participants. The participants provide corrective feedback for each pre-recorded rollout which we then use to train MIND MELD and \textcolor{black}{learn the parameters of MIND MELD's three subnetworks,} $\theta$, $\phi$, and $\phi'$ as well as learn the personalized embedding, $\vec{w}^{(p)}$.

\textcolor{black}{Before providing corrective feedback, participants were given the opportunity to drive in the environment and familiarize themselves \changeNew{with the controls to mitigate learning effects and stabilize performance.}}

\subsection{Ground Truth Data} 
\begin{figure}[t]
    \centering
    \includegraphics[width=.75\columnwidth]{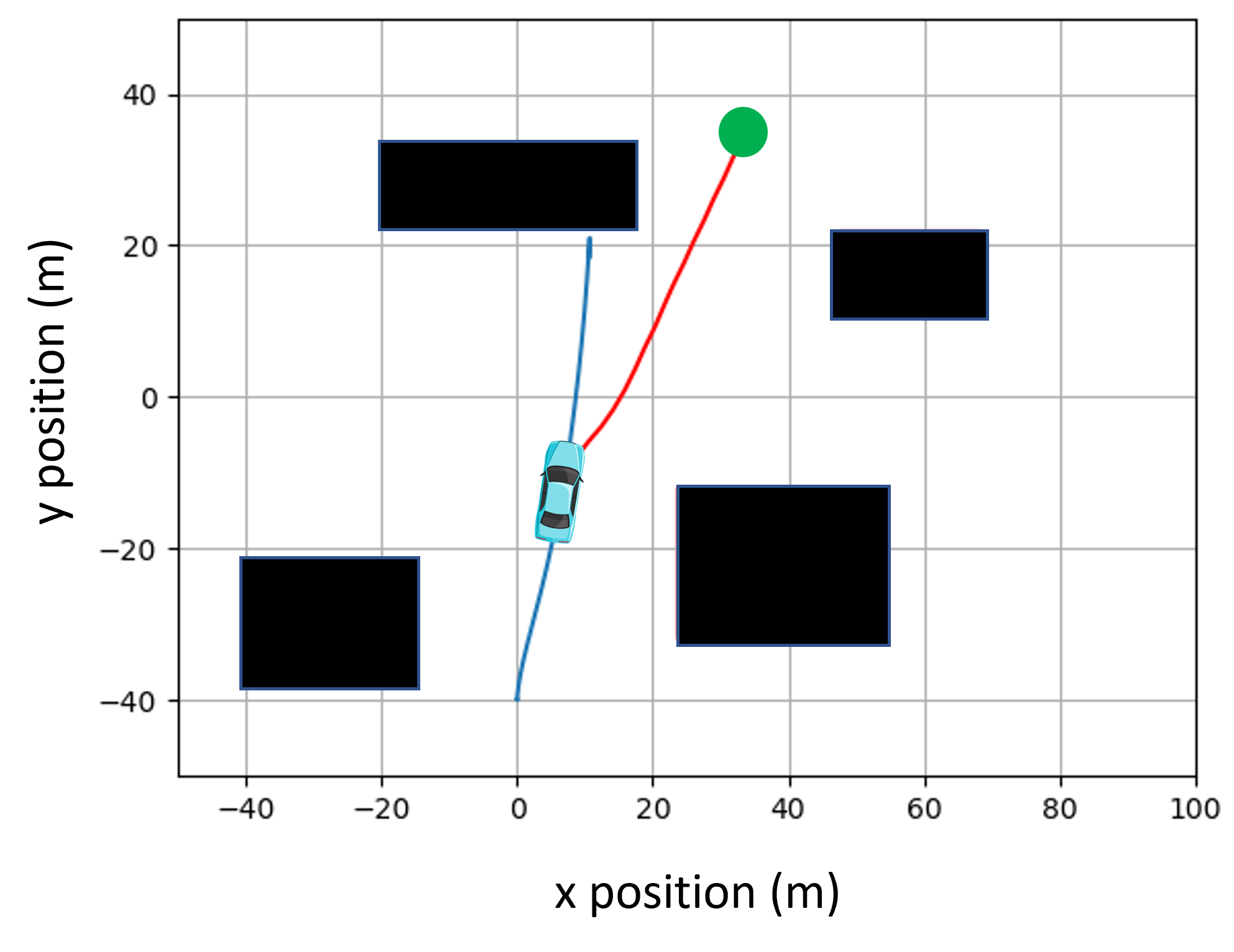}
   \caption{This figure shows the Wizard-of-Oz DAgger rollout in blue and the optimal path produced from RRT* in red at time, $t$. The goal is shown in green and black blocks represent obstacles.}
    \label{fig:rrt}
\end{figure}

To determine ground truth optimal states for the calibration tasks, we employ RRT*  as shown in Fig. \ref{fig:rrt}. At each point along each calibration task trajectory, we determine the optimal path to the goal via RRT*. We then apply a Stanley controller to the path to determine the ground truth label at each point in time.

\subsection{Participants } We recruit 34 training participants via  mailing lists and word of mouth.  Each of these participants provide corrective demonstrations for the calibration tasks which are utilized to train the MIND MELD architecture.

\subsection{Metrics}
Below we discuss the metrics by which we evaluate MIND MELD and the learned embeddings. All survey responses are collected at the beginning of the experiment. All surveys comply with the design guidelines specified in  \citet{Schrum2020FourStudies} and are validated in prior work if possible. 

\textit{Big Five Personality Survey} - We  collect information about the participant's personality via \textcolor{black}{the Mini-IPIP questionnaire \cite{cooper_confirmatory_2010}} to determine if personality correlates with  the learned embedding.

\color{black}\textit{Prior Experience - }
We collect information about a participant's familiarity and experience playing video games, driving a physical car, and driving a virtual car via three Likert scales.  Each Likert scale has 8 items and a 7-point response format (strongly disagree to strongly agree).  We aim to determine whether prior experience with video games or driving correlates with the learned embedding.  

\color{black}

\textit{Trust in Automation -} We measure the participant's trust in automation via the survey presented in \citet{TrustAutomation}. 

\textit{Stylistic tendencies -} We analyzed participants' tendency to either over-correct (i.e., turn the steering wheel too far) or under-correct (i.e., turn the steering wheel not enough) as well as provide delayed or anticipatory feedback via dynamic time warping (DTW) \cite{salvador_fastdtw_2004} between the participant labels, $a$, and ground truth, $o$. \color{black} To estimate the difference in amplitude between the $a$ and $o$ signals, we used DTW to match up the signals in time and calculated the distance, $D_k$, between each time point using the Euclidean distance, $d$. We then summed the distances along the DTW path. To account for whether a participant was over- or under-correcting, we considered whether $a$ or $o$ had a larger magnitude (Eq. \ref{eq:dtw}). 
\begin{align}
D = \sum_{k=0}^K{(-1)^x d(a_k, o_k)} \text{ where}\nonumber \\ x = \begin{cases}
   0 \text{ if } o_k > 0 \text{ and } a_k \geq o_k \\
   1 \text{ if } o_k > 0 \text{ and } a_k < o_k \\
   1 \text{ if } o_k < 0  \text{ and } a_k \geq o_k \\
   0 \text{ if } o_k < 0  \text{ and } a_k < o_k \\
   \end{cases}
    \label{eq:dtw}
\end{align}
 To determine whether a participant was providing delayed or anticipatory feedback, we determined the number of timesteps between $a$ and $o$ on the DTW path. If the majority of points in $a$ were matched to later points in $o$, then our metric for timing was valued as negative or anticipatory; otherwise, our metric was positive or delayed. \color{black}This analysis provided insight into the participants' stylistic tendencies and allowed us to determine if the learned embeddings correlated with these stylistic tendencies.

\subsection{Hypotheses}
\textbf{Hypothesis 1 - } \textit{MIND MELD will improve the corrective labels provided by the participants in the calibration tasks}. We hypothesize that MIND MELD will be able to sufficiently capture the stylistic tendencies of the participants and, based upon these learned embeddings, map the suboptimal labels to labels that more closely approximate the ground truth.

\textbf{Hypothesis 2a - } \textit{The learned embeddings will correlate with personality traits. } We hypothesize that personality traits will inform an individual's corrective style. Therefore, we will find a correlation between the individual's personality and their learned embedding.

\textbf{Hypothesis 2b - } \textit{The learned embeddings will correlate with experience playing video games.} We hypothesize that individuals with more video game experience will \textcolor{black}{provide corrective feedback that more closely approximates optimal feedback,} \changeNew{which would be} reflected in the personalized embedding.

\textbf{Hypothesis 2c - } \textit{The learned embeddings will correlate with driving experience.} We hypothesize that those who often drive cars will find the virtual car and corrective feedback to be counter-intuitive due to the fact that the car does not actually respond to their feedback. Therefore, we expect to see a correlation between experience driving cars and the learned embeddings.

\textbf{Hypothesis 3a - } \textit{The learned embeddings will correlate with participants' tendency to over- or under-correct. } Over- and under-correcting are prominent stylistic tendencies that we observed in pilot participants. Thus, we hypothesize that the learned embeddings will encapsulate an individual's tendency to over- or under-correct and correlate with this tendency.

\textbf{Hypothesis 3b - } \textit{The learned embeddings will correlate with participants' tendency to provide delayed or anticipatory feedback.} We observed that pilot participants tended to either provide feedback that was delayed or anticipatory. Therefore, we hypothesize that the embeddings will correlate with the amount by which the participant provides delayed or anticipatory feedback.

\renewcommand{\arraystretch}{1.3}
\begin{table}[h]
\centering
    \begin{tabular}{|c|c|c|c|}
    \hline
     Survey& Mean&Standard Dev&Cronbach's $\alpha$ \\ \hline\hline 
    Video Games&30.8&11.9&0.90\\ \hline
    Virtual Cars &28.8&10.0&0.91\\ \hline
    Physical Cars&36.7&11.9&0.93\\ \hline
    Trust &64.3&7.0&0.59\\ \hline
    Extraversion &13.3&5.9&0.84\\ \hline
    Agreeableness &20.4&3.8&0.67\\ \hline
    Conscientiousness &20.4&3.6&0.90\\ \hline 
    Neuroticism &9.2&5.5&0.78\\ \hline
    Openness &18.4&5.2&0.80\\ \hline

    \end{tabular}
    \caption{We report average score and Cronbach's $\alpha$ for each survey.}\label{table:descriptive}
 \end{table}

\section{Results}
We recruited 34 participants (Mean age: 28; Standard deviation: 11.9; 29.4\% Female), each of whom completed the calibration tasks and filled out the questionnaires.  The mean score and internal consistency for each questionnaire is reported in Table \ref{table:descriptive}. 
We trained MIND MELD on the participant data and learned each participant's personalized embedding. \changeNew{To train our architecture, we employed Pytorch 3.7 with a learning rate of 0.001.  } We found that we were able to improve participant corrective labels by 63\% on a holdout calibration task. This result supports \textbf{Hypothesis 1} and suggests that MIND MELD was able to learn the stylistic tendencies encapsulated in the embeddings as well as learn how to improve upon suboptimal labels given a participant's learned embedding.

\renewcommand{\arraystretch}{1.3}
\begin{table}[t]
\centering
    \begin{tabular}{|c|c|c|}
    \hline
     Metric &Correlation&p-value  \\ \hline\hline 
    Video Games&0.19&.28\\ \hline
    Virtual Cars &0.24&.18\\ \hline
    Physical Cars&-.25&.15\\ \hline
    Trust &-0.06&.75\\ \hline
    Extraversion &-0.21&.22\\ \hline
    Agreeableness &0.14&.43\\ \hline
    Conscientiousness &-0.033&.89\\ \hline
    Neuroticism &0.25&.16\\ \hline
    Openness &0.17&.34\\ \hline
    \textbf{Over-/Under-Correct} &\textbf{-0.69}&\textbf{$<$.001}\\ \hline
    \textbf{Delay/Anticipate} &\textbf{0.70}&\textbf{$<$.001}\\ \hline

    \end{tabular}
    \caption{We report p-values and correlation coefficients for the correlation analysis between each construct and the learned embeddings. } \label{table:correlation}
 \end{table}

To determine if the learned embeddings correlate with constructs of interest, we performed a correlation analysis between the learned embeddings and participants' experience, personality traits, and stylistic tendencies.  To determine if our data is normally distributed and  homoscedastic, we conducted Shapiro-Wilk's test and the non-constant variance score test, respectively. If the data did not pass the assumption checks, we employed Spearman's test. Otherwise, we used Pearson's test. The results of the correlation analysis between the learned embedding and the construct are reported in Table \ref{table:correlation}. We found participants' stylistic tendencies (i.e., their tendency to over- or under-correct and provide delayed and anticipatory feedback) to be significantly correlated with the learned embeddings. This finding shows that MIND MELD is able to learn a meaningful representation of participants' demonstration styles and supports our \textbf{Hypotheses 3a and 3b}. While we did not find significance at the $\alpha=.05$ level for experience surveys, experience with virtual cars and physical cars are trending towards significance with \textcolor{black}{p-value of .18 and .15,} respectively. \textcolor{black}{Although we do not claim that these results support our hypotheses, since we do not find significance below the $\alpha=.05 $ level, this finding suggests that participants who have more experience operating virtual cars or physical cars may provide stylistically different feedback than those who do not. \color{black} Similarly, we find that the personality trait for neuroticism is trending towards significance (p-value of .16) suggesting that how a person deals with stress may impact their style of providing feedback.} \color{black}We additionally performed a t-test to determine if gender significantly affected the value of the learned embeddings. We found that gender was significant for predicting the learned embeddings with \textcolor{black}{$p=.0015$.}

\changeNew{The relationships we observe between demographic information, stylistic tendencies, and the embedding space suggest that MIND MELD is able to learn meaningful differences between heterogeneous demonstrators. An exploration of what it means to be more ``female" or more of an ``over-corrector" in the context of the embedding space is out of the scope of this paper and, therefore, we do not comment on the meaning of these correlations. Advances in explainable AI may allow us to further explore the meaning behind these correlations in future work.  }

\medskip

\color{black}
\noindent Our main findings are the following:

\begin{itemize}
    \item MIND MELD improves upon suboptimal, human-provided corrective labels by 63\%.
    \item The learned embeddings significantly correlate with demonstrator stylistic tendencies ($p<.001$).
    \item The learned embeddings significantly correlate with demonstrator gender ($p=.0015$).
	\item \textcolor{black}{Participants' experience with virtual cars, ($p=.18$), physical cars ($p=.15$) and neuroticism ($p=.16$) trend towards significance.}
\end{itemize}
\color{black}


\begin{algorithm}[t]

 \begin{algorithmic}[1]
 \small
    \caption{MIND MELD}\label{alg:2}
    \STATE Recruit $M$ training participants
    \FOR{t in training participants}
        \STATE Administer pre-study questionnaires.
      \STATE Collect calibration task data from t.
      \ENDFOR
        \STATE Train MIND MELD on training participant data and learn personalized embeddings $w^{(0:M)}$
        \STATE Freeze architecture parameters, $\phi$, $\phi'$ and $\theta$
        \STATE Recruit test participants
        \FOR{p in test participants}
        \STATE Administer pre-study questionnaires
        \STATE Collect calibration task data from $p$
        \STATE Initialize $\vec{w}^{(p)}$ to $\frac{1}{M}\sum_{i=0}^M \vec{w}^{(i)}$
        \STATE Learn $\vec{w}^{(p)}$via MIND MELD architecture and Eq. \ref{eq:loss}
        \STATE Present LfD algorithm conditions (MIND MELD, BC, and DAgger) in randomized order.
        \FOR{c in conditions}
        \STATE Train agent via condition, $c$, for $N$ demonstrations.
        \ENDFOR
        \STATE Administer post-study questionnaires
        \ENDFOR
     \end{algorithmic}
  \end{algorithm}


\section{Limitations/Future Work} MIND MELD is limited by the fact that training participants are required to learn the personalized embeddings \textcolor{black}{and that we have access to ground truth labels for the calibration tasks.} However, our results demonstrate that MIND MELD improves the quality of the corrective feedback and LfD outcomes, making this additional step a worthwhile endeavor. Additionally, MIND MELD makes several assumptions, as listed in Section \ref{sec:methods}, about the way in which individuals provide corrective feedback. Yet, the success of our algorithm suggests that these assumptions hold for our purposes.

Because MIND MELD requires ground truth labels in the calibration tasks, we are assuming there is one optimal solution to the task (i.e., teach the car the shortest path to the goal).  \changeNew{However, some domains may allow for diverse demonstrations and solutions.}  For example, in autonomous driving, one person might prefer to take the scenic route, while another would prefer the fastest route. 
In this work, we investigate heterogeneity of feedback style, but not the heterogeneity of multiple task solutions.  In future work, we plan to investigate how to account for both the heterogeneity of feedback style and the heterogeneity of task solution preference. 

Furthermore, in future work, we plan to conduct another human-subjects study in which we utilize the MIND MELD architecture to train an LfD agent. This LfD agent will be trained on the improved corrective labels provided by MIND MELD and the policy learned by this agent will be compared to the policies learned via DAgger and BC. To do so, we will recruit a set of test participants who will  first complete the set of calibration tasks which will  be utilized to learn their personalized embeddings. Next, these embeddings will be used to inform the mapping of their feedback when training the learner for the LfD test tasks. Note that we will freeze the parameters $\theta$, $\phi$, and $\phi'$ and we initialize an individual's personalized embedding to the mean of training participants' personalized embeddings.  The steps comprising our study described in this paper and our  future study are \textcolor{black}{shown} in Algorithm 1. 
 
 \textcolor{black}{In future work, we also plan to apply our approach to a robotic pick-and-place task to determine how MIND MELD's abilities to differ in a domain with more degrees of freedom.}


\bibliography{Bibtex.bib}

\begin{thebibliography}{27}
\providecommand{\natexlab}[1]{#1}
\providecommand{\url}[1]{\texttt{#1}}
\providecommand{\urlprefix}{URL }
\expandafter\ifx\csname urlstyle\endcsname\relax
  \providecommand{\doi}[1]{doi:\discretionary{}{}{}#1}\else
  \providecommand{\doi}{doi:\discretionary{}{}{}\begingroup
  \urlstyle{rm}\Url}\fi

\bibitem[{Adams et~al.(2003)Adams, Bruyn, Honde, and
  Angelopoulos}]{TrustAutomation}
Adams, B.~D.; Bruyn, L.~E.; Honde, S.; and Angelopoulos, P. 2003.
\newblock {Trust in automated systems} (June): 136.

\bibitem[{Amershi et~al.(2014)Amershi, Cakmak, Knox, and Kulesza}]{Amershi2014}
Amershi, S.; Cakmak, M.; Knox, W.~B.; and Kulesza, T. 2014.
\newblock {Power to the people: The role of humans in interactive machine
  learning}.
\newblock \emph{AI Magazine} 35(4): 105--120.
\newblock ISSN 07384602.
\newblock \doi{10.1609/aimag.v35i4.2513}.

\bibitem[{Argall et~al.(2009)Argall, Chernova, Veloso, and
  Browning}]{argall_survey_2009}
Argall, B.; Chernova, S.; Veloso, M.~M.; and Browning, B. 2009.
\newblock A survey of robot learning from demonstration.
\newblock \emph{Robotics and Autonomous Systems} 57(5): 469--483.
\newblock \doi{10.1016/j.robot.2008.10.024}.
\newblock \urlprefix\url{https://doi.org/10.1016/j.robot.2008.10.024}.

\bibitem[{Berggren(2019)}]{Berggren2019}
Berggren, J. 2019.
\newblock {Performance Evaluation of Imitation Learning Algorithms with Human
  Experts} .

\bibitem[{Chen et~al.(2016)Chen, Duan, Houthooft, Schulman, Sutskever, and
  Abbeel}]{Chen2016}
Chen, X.; Duan, Y.; Houthooft, R.; Schulman, J.; Sutskever, I.; and Abbeel, P.
  2016.
\newblock {InfoGAN: Interpretable representation learning by information
  maximizing generative adversarial nets}.
\newblock \emph{Advances in Neural Information Processing Systems} 2180--2188.
\newblock ISSN 10495258.

\bibitem[{Chernova and Veloso(2009)}]{Chernova2009}
Chernova, S.; and Veloso, M. 2009.
\newblock {Interactive policy learning through confidence-based autonomy}.
\newblock \emph{Journal of Artificial Intelligence Research} 34: 1--25.
\newblock ISSN 10769757.
\newblock \doi{10.1613/jair.2584}.

\bibitem[{Cooper, Smillie, and Corr(2010)}]{cooper_confirmatory_2010}
Cooper, A.~J.; Smillie, L.~D.; and Corr, P.~J. 2010.
\newblock A confirmatory factor analysis of the {Mini}-{IPIP} five-factor model
  personality scale.
\newblock \emph{Personality and Individual Differences} 48(5): 688--691.
\newblock ISSN 0191-8869.
\newblock \doi{https://doi.org/10.1016/j.paid.2010.01.004}.
\newblock
  \urlprefix\url{https://www.sciencedirect.com/science/article/pii/S019188691000022X}.

\bibitem[{Daume and Eisner(2012)}]{Coaching}
Daume, H.; and Eisner, J. 2012.
\newblock {Imitation Learning by Coaching}.
\newblock \emph{Conference on Neural Information Processing Systems} 1--9.

\bibitem[{Jena, Liu, and Sycara(2020)}]{Jena}
Jena, R.; Liu, C.; and Sycara, K. 2020.
\newblock {Augmenting GAIL with BC for sample efficient imitation learning}
  1--11.

\bibitem[{Kelly et~al.(2019)Kelly, Sidrane, Driggs-Campbell, and
  Kochenderfer}]{HG-Dagger}
Kelly, M.; Sidrane, C.; Driggs-Campbell, K.; and Kochenderfer, M.~J. 2019.
\newblock {HG-DAgger: Interactive imitation learning with human experts}.
\newblock \emph{Proceedings - IEEE International Conference on Robotics and
  Automation} 2019-May: 8077--8083.
\newblock ISSN 10504729.
\newblock \doi{10.1109/ICRA.2019.8793698}.

\bibitem[{Laskey et~al.(2017)Laskey, Chuck, Lee, Mahler, Krishnan, Jamieson,
  Dragan, and Goldberg}]{Laskey2017}
Laskey, M.; Chuck, C.; Lee, J.; Mahler, J.; Krishnan, S.; Jamieson, K.; Dragan,
  A.; and Goldberg, K. 2017.
\newblock {Comparing human-centric and robot-centric sampling for robot deep
  learning from demonstrations}.
\newblock \emph{Proceedings - IEEE International Conference on Robotics and
  Automation} 358--365.
\newblock ISSN 10504729.
\newblock \doi{10.1109/ICRA.2017.7989046}.

\bibitem[{Laskey et~al.(2016)Laskey, Staszak, Hsieh, Mahler, Pokorny, Dragan,
  and Goldberg}]{Laskey2016}
Laskey, M.; Staszak, S.; Hsieh, W. Y.~S.; Mahler, J.; Pokorny, F.~T.; Dragan,
  A.~D.; and Goldberg, K. 2016.
\newblock {SHIV: Reducing supervisor burden in DAgger using support vectors for
  efficient learning from demonstrations in high dimensional state spaces}.
\newblock \emph{Proceedings - IEEE International Conference on Robotics and
  Automation} 2016-June: 462--469.
\newblock ISSN 10504729.
\newblock \doi{10.1109/ICRA.2016.7487167}.

\bibitem[{Liu, Gombolay, and Balakirsky(2021)}]{Liu}
Liu, R.; Gombolay, M.~C.; and Balakirsky, S. 2021.
\newblock {Torwards Unpaired Human-to-Robot Demonstration Translation Learning
  Novel Tasks}.
\newblock \emph{ICSR Workshop Human Robot Interaction for Space Robotics
  (HRI-SR)} .

\bibitem[{Menda, Driggs-Campbell, and Kochenderfer(2019)}]{Menda2019}
Menda, K.; Driggs-Campbell, K.; and Kochenderfer, M.~J. 2019.
\newblock {EnsembleDAgger: A Bayesian Approach to Safe Imitation Learning}.
\newblock \emph{IEEE International Conference on Intelligent Robots and
  Systems} (2): 5041--5048.
\newblock ISSN 21530866.
\newblock \doi{10.1109/IROS40897.2019.8968287}.

\bibitem[{Osa, Neumann, and Peters(2018)}]{Osa2018}
Osa, T.; Neumann, G.; and Peters, J. 2018.
\newblock {An Algorithmic Perspective on Imitation Learning} 7(1): 1--179.
\newblock \doi{10.1561/2300000053}.

\bibitem[{Packard and Onta(2018)}]{Packard2018}
Packard, B.; and Onta, S. 2018.
\newblock {A User Study on Learning from Human Demonstration} (Aiide):
  208--214.

\bibitem[{Paleja and Gombolay(2019)}]{Paleja2019}
Paleja, R.; and Gombolay, M. 2019.
\newblock {Inferring personalized bayesian embeddings for learning from
  heterogeneous demonstration}.
\newblock \emph{arXiv} ISSN 23318422.

\bibitem[{Paleja et~al.(2021)Paleja, Silva, Chen, and
  Gombolay}]{paleja_interpretable_2021}
Paleja, R.; Silva, A.; Chen, L.; and Gombolay, M. 2021.
\newblock Interpretable and Personalized Apprenticeship Scheduling: Learning
  Interpretable Scheduling Policies from Heterogeneous User Demonstrations.

\bibitem[{Ravichandar et~al.(2020)Ravichandar, Polydoros, Chernova, and
  Billard}]{LFDReview}
Ravichandar, H.; Polydoros, A.~S.; Chernova, S.; and Billard, A. 2020.
\newblock \emph{{Recent Advances in Robot Learning from Demonstration}},
  volume~3.
\newblock ISBN 1008190632.
\newblock \doi{10.1146/annurev-control-100819-063206}.

\bibitem[{Ross and Bagnell(2010)}]{Ross2010}
Ross, S.; and Bagnell, J.~A. 2010.
\newblock {Efficient reductions for imitation learning}.
\newblock \emph{Journal of Machine Learning Research} 9: 661--668.
\newblock ISSN 15324435.

\bibitem[{Ross, Gordon, and Bagnell(2011)}]{Ross2011}
Ross, S.; Gordon, G.~J.; and Bagnell, J.~A. 2011.
\newblock {No-regret reductions for imitation learning and structured
  prediction}.
\newblock \emph{Aistats} 15: 627--635.
\newblock \urlprefix\url{http://proceedings.mlr.press/v15/ross11a/ross11a.pdf}.

\bibitem[{Ross et~al.(2013)Ross, Melik-Barkhudarov, Shankar, Wendel, Dey,
  Bagnell, and Hebert}]{ross_learning_2013}
Ross, S.; Melik-Barkhudarov, N.; Shankar, K.~S.; Wendel, A.; Dey, D.; Bagnell,
  J.~A.; and Hebert, M. 2013.
\newblock Learning monocular reactive {UAV} control in cluttered natural
  environments.
\newblock In \emph{2013 {IEEE} {International} {Conference} on {Robotics} and
  {Automation}}, 1765--1772.
\newblock \doi{10.1109/ICRA.2013.6630809}.

\bibitem[{Salvador and Chan(2004)}]{salvador_fastdtw_2004}
Salvador, S.; and Chan, P. 2004.
\newblock {FastDTW}: {Toward} {Accurate} {Dynamic} {Time} {Warping} in {Linear}
  {Time} and {Space}.

\bibitem[{Sammut(1992)}]{Sammut1992}
Sammut, C. 1992.
\newblock {Automatically Constructing Control Systems by Observing Human
  Behaviour}.
\newblock \emph{Second International Inductive Logic Programming Workshop}
  (May).

\bibitem[{Schrum et~al.(2020)Schrum, Johnson, Ghuy, and
  Gombolay}]{Schrum2020FourStudies}
Schrum, M.~L.; Johnson, M.; Ghuy, M.; and Gombolay, M.~C. 2020.
\newblock {Four years in review: Statistical practices of likert scales in
  human-robot interaction studies}.
\newblock \emph{ACM/IEEE International Conference on Human-Robot Interaction}
  43--52.
\newblock ISSN 21672148.
\newblock \doi{10.1145/3371382.3380739}.

\bibitem[{Sena and Howard(2020)}]{Sena2020}
Sena, A.; and Howard, M. 2020.
\newblock {Quantifying teaching behavior in robot learning from demonstration}
  \doi{10.1177/0278364919884623}.

\bibitem[{Spencer et~al.(2020)Spencer, Choudhury, Barnes, Schmittle, Chiang,
  Ramadge, and Srinivasa}]{Spencer2020}
Spencer, J.; Choudhury, S.; Barnes, M.; Schmittle, M.; Chiang, M.; Ramadge, P.;
  and Srinivasa, S. 2020.
\newblock {Learning from Interventions: Human-robot interaction as both
  explicit and implicit feedback} \doi{10.15607/rss.2020.xvi.055}.

\end{thebibliography}

\end{document}